\documentclass[a4paper]{svmult}
\usepackage{imakeidx}
\usepackage[utf8]{inputenc}

\usepackage{url}

\usepackage{type1cm}        
\usepackage{makeidx}         
\usepackage{graphicx}        
\usepackage{multicol}        
\usepackage[bottom]{footmisc}

\usepackage{newtxtext}       %
\usepackage{newtxmath}       
\usepackage{color,xcolor,ucs}
\usepackage{subfig}
\usepackage{floatrow}
\usepackage{tabularx}
\usepackage{float}
\usepackage{amsmath}
\usepackage{amssymb}  

\usepackage{cite}
\usepackage{xr-hyper}
\usepackage{wrapfig}

\usepackage{algorithm}
\usepackage{algpseudocode}
\usepackage{multirow}
\usepackage{listings}
\usepackage{lipsum}
\usepackage{textcomp} 
\makeatletter
\algnewcommand{\LineComment}[1]{\Statex \hskip\ALG@thistlm \(\triangleright\) #1}
\algnewcommand{\LineCommentCont}[1]{\Statex \hskip\ALG@thistlm \parbox[t]{\linegoal}{\hangindent=1em\hangafter=1 $\triangleright$ #1}}
\makeatother
\makeindex[options=-s svindd.ist]            

\begin{document}

\title*{Task-Motion Planning for Navigation in Belief Space}
\author{Antony Thomas \and Fulvio Mastrogiovanni
\and Marco Baglietto}
\institute{The authors are with the Department of Informatics, Bioengineering, Robotics, and Systems Engineering, University of Genoa, Via All'Opera Pia 13, 16145 Genoa, Italy. \\\email{antony.thomas@dibris.unige.it, {fulvio.mastrogiovanni, marco.baglietto}@unige.it}}
%
%
\maketitle

\abstract*{We present an integrated Task-Motion Planning (TMP) framework for navigation in large-scale  environment. Autonomous robots operating in real world complex scenarios require planning in the discrete (task) space and the continuous (motion) space. In knowledge intensive domains, on the one hand, a robot has to reason at the highest-level, for example the regions to navigate to; on the other hand, the feasibility of the respective navigation tasks have to be checked at the execution level. This presents a need for motion-planning-aware task planners. We discuss a probabilistically complete approach that leverages this task-motion interaction for navigating in indoor domains, returning a plan that is optimal at the task-level. Furthermore, our framework is intended for motion planning under motion and sensing uncertainty, which is formally known as belief space planning. The underlying methodology is validated with a simulated office environment in Gazebo. In addition, we discuss the limitations and provide suggestions for improvements and future work.}

\abstract{We present an integrated Task-Motion Planning (TMP) framework for navigation in large-scale  environment. Autonomous robots operating in real world complex scenarios require planning in the discrete (task) space and the continuous (motion) space. In knowledge intensive domains, on the one hand, a robot has to reason at the highest-level, for example the regions to navigate to; on the other hand, the feasibility of the respective navigation tasks have to be checked at the execution level. This presents a need for motion-planning-aware task planners. We discuss a probabilistically complete approach that leverages this task-motion interaction for navigating in indoor domains, returning a plan that is optimal at the task-level. Furthermore, our framework is intended for motion planning under motion and sensing uncertainty, which is formally known as belief space planning. The underlying methodology is validated with a simulated office environment in Gazebo. In addition, we discuss the limitations and provide suggestions for improvements and future work.}

\section{Introduction}
Autonomous robots operating in complex real world scenarios require different levels of planning to execute their tasks. High-level (task) planning helps break down a given set of tasks into a sequence of sub-tasks. Actual execution of each of these sub-tasks would require low-level control actions to generate appropriate robot motions. In fact, the dependency between logical and geometrical aspects is pervasive in both task planning and execution. Hence, planning should be performed in the task-motion or the discrete-continuous space. 

In recent years, combining high-level task planning with low-level motion planning has been a subject of great interest among the Robotics and Artificial Intelligence (AI) community. Traditionally, task planning and motion planning have evolved as two independent fields. AI planning frameworks as the Planning Domain Definition Language (PDDL)~\cite{mcdermott1998AIPS} mainly focus on high-level task planning supposing that the geometric preconditions (e.g., grasping poses for a pick-up task~\cite{srivastava2014ICRA}) for the robot motion to carry out these tasks are achievable. However, in reality, such an assumption can be catastrophic as an action or sequence of actions generated by the task planner might turn out to be unfeasible at the controller execution level. 

This paper contributes an approach that provides an interface between task and motion planning for navigating in large knowledge-intensive domains. Such domains require a robot to reason about different objects and locations to navigate to, subject to expending as less cost as possible. Our task-motion interface layer facilitates this reasoning by communicating the motion feasibility and the corresponding planned motion costs to the task planner, synthesizing an optimal plan. While our approach is applicable to any domains that require task-motion interaction, we establish the key ideas through a motivating example.


\begin{wrapfigure}{r}{0.4\textwidth}
\vspace{-24pt}
  \begin{center}
    \includegraphics[width=0.75\textwidth]{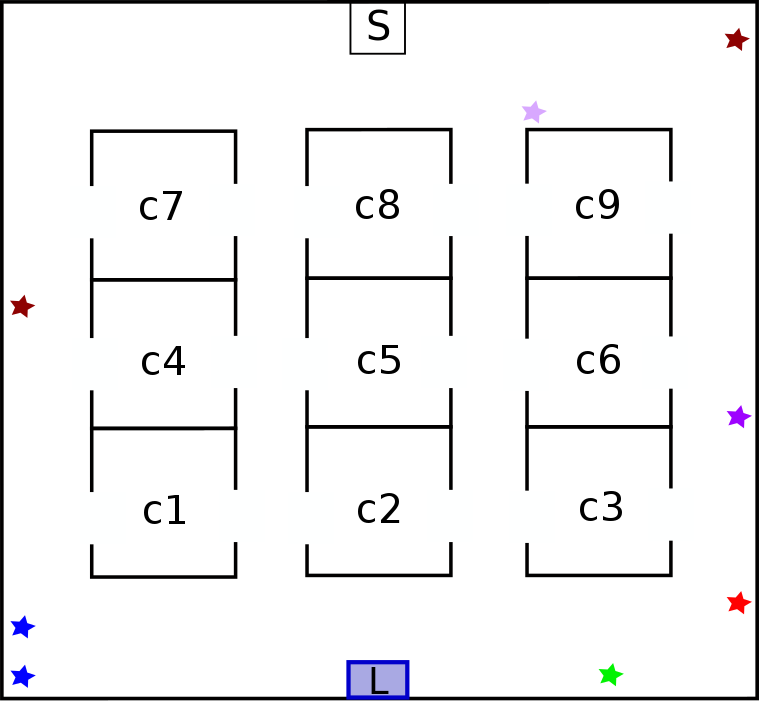}
  \end{center}
    \caption{Blueprint of an office environment.}
    \label{fig:office}
    \vspace{-5pt}
     \vspace{-5pt}
\end{wrapfigure}


\textit{Motivating example:} Consider the office environment shown in Fig~\ref{fig:office}. The regions $\{c_i\}_{(i=1,...,9)}$ are cubicles and $L$ denotes a lift. The robot, starting from region $S$ have to visit certain cubicles to receive documents. These documents then have to be delivered to the next floor, via the lift $L$. The stars with different colors represent certain unique features like, printer, trash can, lounge, that aids the robot in better localization. Once the robot knows the regions to visit, then it suffices to perform $goto$ $c_i$ actions and collect the documents from these regions. However, to synthesize an optimal plan it is necessary to sequence these actions in an order that minimizes the cost function. It is therefore inevitable to obtain the motion costs of these $goto$ $c_i$ actions, so as to accurately synthesize the optimal plan.

Yet, real-world scenarios often induce uncertainties. Such uncertainties arise due to insufficient knowledge about the environment, inexact robot motion or imperfect sensing. In such scenarios, the robot poses or other variables of interest can only be dealt with, in terms of probabilities. Planning is therefore done in the \textit{belief} space, which corresponds to the probability distributions over possible robot states. Consequently, for efficient planning and decision making, it is required to reason about future belief distributions due to candidate actions and the corresponding expected observations. Such a problem falls under the category of Partially Observable Markov Decision Processes (POMDPs)~\cite{kaelbling1998AI}. Our motion planner is therefore equipped to perform planning in partially-observable state-spaces with motion and sensing uncertainty.

\begin{figure}[t]
	\centering
		\includegraphics[scale=0.10]{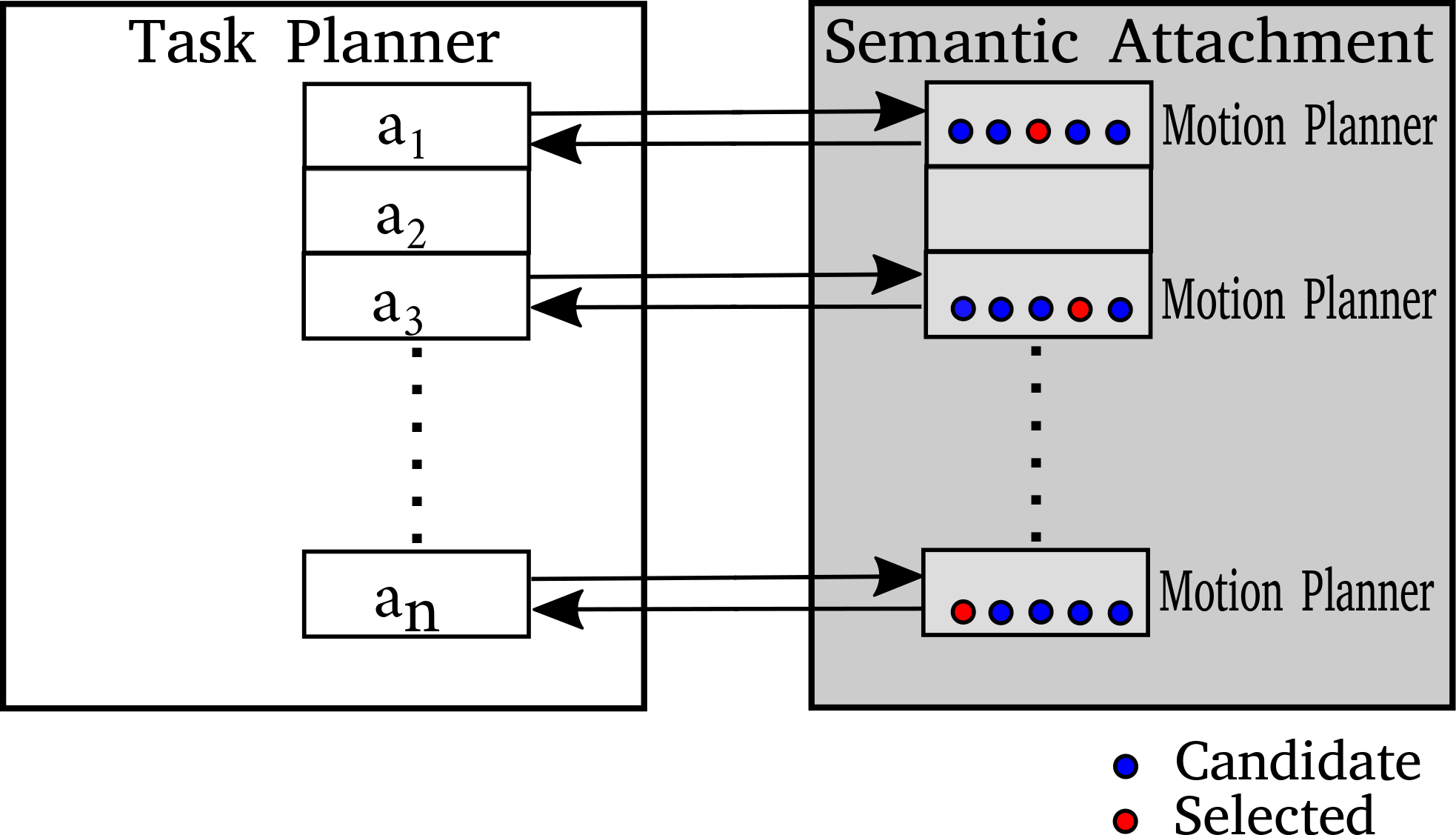}
		\caption{$a_1$, $a_2$, $a_3$, ..., $a_n$ are the discrete actions. Different motion plans are generated for the symbolic actions via semantic attachment. The best candidate (red in figure) among these is then selected, returning the motion cost to the task planner.}
	\label{fig:sofar}
\end{figure}

We develop a probabilistically complete Task-Motion Planning (TMP) framework for mobile robot navigation under partial-observability, embedding a motion planner within a task planner through an interface layer. An overview of our TMP approach is shown in Fig.~\ref{fig:sofar}. $A = \{a_1,...,a_n\}$ is the finite set of symbolic actions available to the task planner. Once an action that require appropriate robot motions to be generated is expanded by the task planner, a call to an external library is triggered. The symbolic parameters are then converted to their corresponding geometric instantiations. For example, for an action that takes the robot to a particular region, the instantiations would be the different sampled poses in that region. These geometric instantiations are pre-sampled since the map of the environment is known. The instantiations give rise to different motion plans and the best among them is chosen according to a certain metric. The cost of the selected motion plan is then returned to the task planner as the cost of the corresponding action. 

\label{sec:one}

\section{Related Work}

TMP has emerged as an active research area in the recent past, with particular focus on manipulation robots. Manipulation task are often rendered infeasible due to the robot's end-effector's reachability workspace. This calls for an integrated TMP approach to ensure geometric feasibility of the high-level tasks.   

The genesis of TMP can be credited to Fikes and Nilsson for their work on STRIPS~\cite{fikes1971strips} which further led to the Shakey project~\cite{nilsson1984shakey}. Shakey's planner performed a logical search first, assuming that the resulting robot motion plans can be formulated. This assumption limits the capability of the agent as the high-level actions may turn out to be non executable due to geometric limitations. Later works either carried out the generated plans, validating them using a robot motion planner~\cite{dornhege2009SSRR} or performed a combined search in the logical and geometric spaces using a state composed of both the symbolic and geometric paths~\cite{cambon2009IJRR}. The aSyMov planner in~\cite{cambon2009IJRR} adopts a combination of Metric-FF~\cite{hoffmann2003JAIR} and a sampling based motion planner. In contrast, we use a temporal task planner, POPF-TIF~\cite{piacentini2015AI} with roadmap based sampling, incorporating robot state uncertainty. Srivastava \textit{et al.}~\cite{srivastava2014ICRA} implicitly incorporate geometric variables, performing symbolic-geometric mapping using a planner-independent interface layer.

Kaelbling and Lozano-P\'{e}res~\cite{kaelbling2012aTR} propose a hierarchical approach that tightly integrates the logical and geometric planning. The complexities arising out of long horizon planning are tackled to the extent that planning is done at different levels of abstraction, thereby reducing the long horizons to a number of feasible sub-plans of shorter horizon. This regression-based planner assumes that the actions are reversible while backtracking. In contrast to their earlier work the serializability assumption of the subgoals is relaxed. This work is extended in~\cite{kaelbling2013IJRR} to consider the current state uncertainty, modeling the planning problem in the belief space. Uncertain outcomes are modeled by converting  a Markov decision processes (MDP) into a weighted graph, thereby modifying their earlier approach of \textit{hierarchical planning in the now}. Belief update is then performed when observations are obtained. FFRob~\cite{garrett2018IJRR} performs task planning, by performing search over a sampled finite set of poses, grasps and configurations. They extend the FF heuristics, incorporating geometric and kinematic
planning constraints that provide a tight estimate of the distance to the goal. 



Toussaint~\cite{toussaint2015IJCAI} performs optimization over an objective function based on the final geometric configuration (and the cost thereby), finding approximately locally optimal solutions by minimizing the objective function. The planning problem is modeled as a constraint satisfaction problem with symbolic states used to define the constraints in the optimization. Lozano-P\'{e}res and Kaelbling~\cite{lozano2014IROS} model the motion planning as a constraint satisfaction problem over a subset of the configuration space. Iteratively Deepened Task and Motion Planning (IDTMP) is a constraint based task planning approach that incorporates geometric information (motion feasibility) at the task planning level~\cite{dantam2018IJRR}. In our architecture, the motion costs are returned to the task planner, similar to the motion planner information that guides the IDTMP task planner. IDTMP performs task-motion interaction using abstraction and refinement functions whereas we use \textit{semantic attachments}~\cite{dornhege2009ICAPS} to that aim.

Though the above discussed approaches fall under the category of TMP for manipulation, the scope of TMP is not limit to manipulation problems alone. TMP for navigation is pervasive in most real world scenarios. Yet, TMP for robot navigation has received less attention in the past. While navigating in large scale environments it is straightforward to plan first in terms of the places to navigate before steering the robot (see \textit{motivating example}, Section~\ref{sec:one}). This calls for task plans that are motion-plan aware, in terms of motion costs and its feasibility. PETLON~\cite{lo2018AAMAS} is the closest work to our approach since they also discuss a TMP approach for navigation that is task level optimal. However, the action costs returned by their motion planner is the trajectory length and they assume completely observable domains. In contrast, our framework is more general, since our motion planner assumes partial observability by performing planning in the belief space. In Section~\ref{sec:results}, we benchmark the scalability of our approach by comparing with a motion planner that evaluates the geometric-level cost of navigation.

\section{Preliminaries and Definitions}
TMP essentially involves combining discrete and continuous decision-making to facilitate efficient interaction between the two domains. Below we define the TMP problem formally. The notations and formalism correspond to that of a state-transition system~\cite{ghallab2004automated}.

\begin{definition}\textit{Task} domain can be represented via state transition system and is a tuple $\Omega = (S, A, \gamma, s_0, S_g)$ where
\end{definition}
\begin{itemize}
\item $S$ is a finite set of states, each state is a conjunction of propositions.
\item $A$ is a finite set of actions
\item $\gamma : S \times A \rightarrow S$ is the state transition function such that $s' = \gamma(s, a)$
\item $s_0 \in S$ is the start state
\item $S_g \subseteq S$ is the set of goal states
\end{itemize}

\begin{definition}\textit{Task} Plan for a task domain $\Omega$ is the sequence of actions $a_0,...,a_n$ such that $s_{i+1} = \gamma(s_i, a_i)$, for $i = 0,...,n$ and $s_{n+1}$ satisfies $S_g$.
\end{definition}
 
Due to the popularity of PDDL among the Planning community, we resort to the same for modeling our task domain. PDDL is an action-centred language, where each action, $a_i$ is described as a tuple $a_i = (pre_{a_i},eff_{a_i})$. $pre_{a_i}$ (precondition for $a_i$ ) is a conjunction of positive and negative propositions that must hold for action execution and $eff_{a_i}$ (effects of $a_i$) is a conjunction of positive ($eff^+_{a_i}$) and negative ($eff^-_{a_i}$) propositions that are added or deleted upon action application, thereby changing the system state. The positive effects, $eff^+_{a_i}$, is the set of propositions that become true upon the execution of action $a_i$ and the negative effects, $eff^-_{a_i}$, is the set of propositions that evaluates to false. An action $a_i$ is said to be applicable to a state $s_i$ if each proposition of the preconditions hold in $s_i$, that is, $pre_a \subseteq s_i$.  If an action $a_i$ is applicable in state $s_i$, the corresponding successor state $s_{i+1}$ is obtained as, $s_{i+1} = \gamma(s_i, a_i)$, where $s_{i+1} = (s_i \setminus eff^-_{a_i}) \cup eff^+_{a_i}$. A valid plan is a sequence of actions that when executed from $s_0$, results in $S_g$. 

A planning problem with PDDL is created by providing a domain description- that describes the predicates and action schemas with free variables, and a problem description- that specifies the objects, initial state and the goal condition. The objects are used to instantiate the predicates and action schemas, through a process called grounding. Grounding is the process by which every combination of objects is used to replace the free variables in predicates and action schemas to obtain propositions and ground actions respectively. In this paper, we use an extension of PDDL~\cite{fox2003JAIR} that supports durative actions and numeric-valued fluents. A task domain that supports the same, can be written as, $\Omega = (S, V, A, \gamma, s_0, S_g)$, where $V$ is a set of real valued variables, $s_0 \in S \cup V$ and $S_g \subseteq S \cup V$. A durative action is a tuple  $a_i = (pre_{a_i},eff_{a_i},dur_{a_i})$, where $pre_{a_i}$ and $eff_{a_i}$ are temporally annotated by specifying conditions/effects that holds at the \textit{start}, \textit{end} or during the entire interval and are expressed using the constructs \textit{at start}, \textit{at end} and \textit{over all} respectively. $dur_{a_i}$ corresponds to the duration of action $a_i$.

\begin{definition}\textit{Motion} Planning Problem is a tuple $M = (C, f, q_0, G)$ where
\end{definition}
\begin{itemize}
\item $C$ is the configuration space or the space of possible robot poses
\item $f =\{1,0\}$ determines if a configuration is collision free ($C_{free}$ with $f =1$) or not ($f=0$)  
\item $q_0$ is the initial configuration
\item $G$ is the set of goal configurations
\end{itemize}

A motion plan essentially involves finding a valid trajectory in $C$ from $q_0$ to $q_n \in G$ such that $f$ evaluates to true for $q_0,...,q_n$. A motion plan can also be defined as $\tau : [0, 1] \rightarrow C_{free}$ such that $\tau(0) = q_0$ and $\tau(1) \in G$. We will use a combination of the two to define the TMP problem and use roadmap based motion planner to generate collision free configurations.  

\begin{definition}\textit{Task-Motion} Planning Problem is a tuple $\Psi =(C, \Omega, \phi, \xi)$ where
\end{definition}
\begin{itemize}
\item $\phi : S  \rightarrow 2^ C$, mapping states to the configuration space 
\item $\xi : A  \rightarrow 2^ C$, mapping actions to motion plans
\end{itemize}
and the TMP problem is to find a sequence of actions $a_0,...,a_n$ such that $s_{i+1} = \gamma(s_i, a_i)$, $s_{n+1} \in S_g$ and to find a sequence of motion plans $\tau_0,...,\tau_n$, $\tau_n(1) \in G$ such that for $i = 0,...,n$, it holds that

\vspace{-0.5cm}

\begin{align}
& \tau_i(0) \in \phi(s_i) \ \textrm{and} \ \tau_i(1)  \in \phi(s_{i+1})  \\
&\tau_{i+1}(0) = \tau_i(1)   \\
&\tau_i \in \xi(a_i)
\end{align}

In this paper, we consider the TMP problem for a mobile robot operating in a partially-observable, pre-mapped environment.
At any time $k$, we denote the robot pose (or configuration $q_k$) by $x_k\doteq(x, y, \theta)$, the measurement acquired is denoted by $z_k$ and the control action applied is denoted as $u_k$. We consider a standard motion model with Gaussian noise



\vspace{-0.40cm}
\begin{equation}
x_{k+1} = f(x_k,u_k)+w_k \  ,  \ w_k \sim \mathcal{N}(0,R_k)
\label{eq:odometry_model}
\end{equation}

\noindent

where $w_k$ and $R_k$ are the Gaussian noise and process covariance respectively. To process the landmarks in the environment we measure the range and the bearing of each landmark relative to the robot's local coordinate frame. In general, we consider the observation model with Gaussian noise, 

\vspace{-0.35cm}
\begin{equation}
z_k = h(x_k) + v_k \  ,  \ v_k \sim \mathcal{N}(0,Q_k)
\label{eq:measurement_model}
\end{equation}
 

It is to be noted that we assume data association as solved and hence given a measurement we know the corresponding landmark that generated it. The motion (\ref{eq:odometry_model}) and observation (\ref{eq:measurement_model}) models can be written probabilistically as
$p(x_{k+1}|x_k, u_k)$ and $p(z_k|x_k)$ respectively. Given an initial distribution $p(x_0)$, and the motion and observation models, the posterior probability distribution at time $k$ can be written as

\vspace{-0.2cm}
\begin{equation}
p(X_{0:k}|Z_{0:k},U_{0:k-1}) = p(x_0)\prod_{i=1}^k p(x_{k}|x_{k-1}, u_{k-1}) p(z_k|x_k)
\end{equation}
where $X_{0:k} \doteq \{x_0,...,x_k\}$, $Z_{0:k}  \doteq\{z_0,...,z_k\}$ and $U_{0:k-1} \doteq \{u_0,...,u_{k-1}\}$. This posterior probability distribution is the \textit{belief} at time $k$, denoted by $b[X_k] \sim \mathcal{N} (\mu_k, \Sigma_k)$. Similarly, given an action $u_k$, the propagated belief can be written as

\vspace{-0.3cm}
\begin{equation}
b[\bar{X_{k+1}}] = p(X_k|Z_k, U_{k-1})p(x_{k+1}|x_k, u_k)
\end{equation}

Given the current belief $b[X_k]$, the control $u_k$, the propagated belief parameters can be computed using the standard Extended Kalman Filter (EKF) prediction as 

\vspace{-0.5cm}
\begin{equation}
\begin{split}
\bar{\mu}_{k+1} & = f(\mu_k, u_k)\\
\bar{\Sigma}_{k+1}   & = F_{k} \Sigma_k F_{k}^T + R_k
\end{split}
\label{eq:predict}
\end{equation}
where $F_k$ is the Jacobian of $f(\cdot)$ with respect to $x_k$. Upon receiving a measurement $z_k$, the posterior belief $b[X_{k+1}]$ is computed using the EKF update equations

\vspace{-0.4cm}
\begin{equation}
\begin{split}
K_k     & = \bar{\Sigma}_{k+1} H_k^T(H_k \bar{\Sigma}_{k+1}  H_k^T + Q_k)^{-1}\\
\mu_{k+1} & = \bar{\mu}_{k+1} + K_k(z_{k+1}-h(\bar{\mu}_{k+1},l_i))\\
\Sigma_{k+1} & = (I -K_k H_k)\bar{\Sigma}_{k+1} 
\end{split}
\label{eq:update}
\end{equation}

\noindent
where $H_k$ is the Jacobian of $h(\cdot)$ with respect to $x_k$, $K_k$ is the Kalman gain and $I \in \mathbb{R}^{3 \times 3}$.

\section{Approach}
PDDL based planning frameworks are limited, as they are incapable of handling rigorous numerical calculations. Most approaches perform such calculations via external modules or \textit{semantic attachments}, e.g.~\cite{dornhege2009ICAPS}. The term semantic attachment was coined by Weyhrauch~\cite{weyhrauch1980AI} to describe attaching algorithms to function and predicate symbols via external procedures. However, the effects returned by these semantic attachments are not exploited in identifying \textit{helpful actions} during search and hence do not provide any heuristic guidance, deeming the task unsolvable most often. An action is considered \textit{helpful} if it achieves at least one of the lowest level goals in the relaxed plan to the state at hand~\cite{hoffmann2003JAIR}. Recently, Bernardini \textit{et al.}~\cite{bernardini2017ICAPS} developed a PDDL based temporal planner to implicitly trigger such external calls via a specialized semantic attachments called \textit{external advisors}. They classify variables into direct ($V^{dir}$), indirect ($V^{ind}$) and free ($V^{free}$). $V^{dir}$/$V^{free}$ variables are the normal PDDL function variables whose values are changed in the action effects, in accordance with PDDL semantics. $V^{ind}$ variables are affected by the changes in the $V^{dir}$ variables. A change in a $V^{dir}$ variable invokes the external advisor which in turn computes the $V^{ind}$ variables. The Temporal Relaxed Plan Graph (TRPG)~\cite{coles2010ICAPS} construction stage of the planner incorporates the indirect variable values for heuristic calculation, thereby synthesizing an efficient goal-directed search. We employ this semantic attachment based approach for task-motion interaction. The overall procedure and the interface layer are discussed in detail in the remainder of this section. 


\vspace{-0.4cm}
\subsection{Task Planner} 

TMP for navigation requires that the task planner takes into account the motion feasibility and the corresponding motion costs while synthesizing a plan. As opposed to the manipulation domain, where the motion feasibility is corroborated with the end-effector's reachability workspace, in navigation domains this is often validated against the cost constraints. For example, a robot navigating in a corridor with a bound on the pose covariance to avoid collision. In our tests, PDDL is used to define the task domain.

A fragment of the PDDL domain is shown in Fig.~\ref{fig:domain}. The actions \textit{goto\_region} and \textit{goto\_lift} invoke the external module call once the facts \textit{(increase (act-cost) (external))} and \textit{(increase (goal-trace) (bound))} is encountered. Here, \textit{act-cost} and \textit{goal-trace} are the $V^{dir}$ variables and \textit{external} and \textit{bound} are the $V^{ind}$ variables. The function \textit{(triggered ?from ?to)} is assigned unity each time the actions are expanded and re-initialized to zero once the action duration is completed. In this way, the grounded variables \textit{from} (start) and \textit{to} (goal) are communicated to the motion planner. The variables \textit{extern} and \textit{bound} returns the motion cost and goal covariance trace respectively, which are computed by the external module. The action \textit{collect\_document} does not invoke the motion planner. The function \textit{get ?r}, where \textit{r} is a free variable denoting cubicles, is initialized to unity for the cubicles from which the documents are to be collected and to zero for the remaining. 
\vspace{-0.4cm}
\subsection{Motion Planner} 
We use a sampling based Probabilistic Roadmap (PRM)~\cite{kavraki1996IEEE} to instantiate robot poses for the task actions. To begin with, the initial mean and covariance of the robot pose is assumed to be known. This means that the initial state $s_0$ corresponds to a single pose instantiation $q_0$. The regions to be navigated to (as discussed in the \textit{motivating example}) are also instantiated into poses, by sampling from the pose space within each region/cubicle. In other words, the pose instantiations are the poses that lie inside cubicles and are sampled once the map of the environment is available.  Once an action $a_i$ (\textit{goto\_region} or \textit{goto\_lift}) is expanded by the task planner, the corresponding start and goal state, that is $s_i$ and $s_{i+1}$ are communicated to the motion planner. With the pose instantiation of $s_i$ as the start node, for each pose instantiation of $s_{i+1}$, a breadth first search is performed, parallelly simulating a sequence of controls long each edge and estimating the beliefs at the nodes. The instantiation corresponding to the minimum cost is then selected as the goal pose to be arrived at, for the state $s_{i+1}$. This instantiation then becomes the start node when an expansion is attempted from state $s_{i+1}$.

Since we plan in the belief space of the robot state, given the mean and covariance of the starting node (initial mean pose and covariance known) we propagate the belief along the edges of the PRM as the roadmap is expanded following a breadth first search. Belief update is performed upon reaching a node if a landmark falls within the sensor range. Since we are in the planning phase and yet to obtain observations, we simulate future observations $z_{k+1}$ given the propagated belief $b[\bar{X_{k+1}}]$, the set of landmarks $lm$ and the measurement model (\ref{eq:measurement_model}). Given a pose $x \in b[\bar{X_{k+1}}]$, the nominal observation $\hat{z} = h(x, lm_i)$ is corrupted with noise to obtain $z_{k+1}$. 
 
\vspace{-0.5cm}
\subsection{Task-Motion Planning for Navigation}

\begin{algorithm}[t!]
\label{alg:problem statement}
\caption{TMP for Navigation in Belief Space}
\label{algo:TMP}
\begin{algorithmic}[1]
\Require{$\Omega$, $M$, $\phi$, $s_0$, $\xi$, $S_g$, $q_0$ , $\eta$}
\While{true}
\State{$a_i$ $\leftarrow$ TRPG construction of Task Planner}
\LineComment $a_i$ = an action selected to expand to the next symbolic state
\If{$a_i$ $\in$ \{\textit{goto\_region, \textit{goto\_lift}}\}}
\State{External module $\leftarrow$ $V^{dir}$}
\LineComment $V^{dir}$  $= \{$\texttt{act-cost}, \texttt{goal-trace}$\}$
\State{$s_i = from$, $s_{i+1} = to$}
\State {cost $\leftarrow$ $\emptyset$, motion\_plan $\leftarrow$ $\emptyset$ }

 \For{j=1:p}
\State{BFS with belief propagation along the edges}
\LineComment start node $=\tau_i^j(0) = \phi(s_i)$, goal node $=\tau_i^j(1) = \phi(s_{i+1})$

\State{cost.push($c_{\Sigma}^j$), motion\_plan.push($\tau_i^j$) }
   \EndFor

\State {\texttt{external}  $\leftarrow$ \textbf{min} cost, $j^*$ = \textbf{arg\,min} cost }
\State {\texttt{bound} $\leftarrow$ $trace(\Sigma_{\tau_i^{j^*}(1)})$}
\State {$\tau_i \leftarrow \tau_i^{j^*}$}

\EndIf

\EndWhile
\State {$\pi^*$ $\leftarrow$ Task Planner($\Psi$)}
 \For{each $a_i \in \pi^*$ }
\If{\texttt{bound} > $\eta$}
\State{$\pi^*$ $\leftarrow$ $\emptyset$}
\State {break}
\EndIf
   \EndFor
\State \Return{$\pi^*$}
\end{algorithmic}
\end{algorithm}

An overview of our TMP approach is presented in Algorithm~\ref{algo:TMP}. In our approach, the task-motion interaction occurs through semantic attachment by dynamically loading a shared library, that is passed to the planner as an input. The semantic attachment is called only if the action to be expanded is either \textit{goto\_region} or \textit{goto\_lift} (see line 3). Once an action $a_i$ is expanded by the task planner, the corresponding start ($s_i$) and goal ($s_{i+1}$) state are communicated to the motion planner through the the function \textit{(triggered ?from ?to)} (see line 5). The pose instantiation of $s_i$, that is, $\tau_i(0) = \phi(s_i)$ is known, since it is the mean of the
current belief distribution. For each pose instantiation of $\tau_i^j(1) = \phi(s_{i+1})$, a motion plan is attempted with $\tau_i(0) = \tau_i^1(0)=...=\tau_i^p(0)$ as the start node, where $p$ is the number of pose instantiations of $s_{i+1}$. The set of feasible motion plans is obtained by performing a breadth first search over the roadmap. Along each edge of the roadmap, the belief at $s_i$ is propagated using by simulating the sequence of controls and observations. We use EKF to compute the appropriate matrices for belief computation. Posterior belief is computed at each node if a landmark falls within the sensor range. From the set of feasible motion plans generated, the minimum cost and the corresponding goal state trace, $trace(\Sigma_{\tau_i^{j^*}(1)})$, are assigned to the $V^{ind}$ variables, \textit{external} and \textit{bound} (see lines 11-12). The corresponding motion plan ($\tau_i$) and the goal node ($\tau_i^{j^*}(1)$) are also stored. The goal node $\tau_i^{j^*}(1)$, subsequently becomes the start node for the roadmap search from $s_{i+1}$. Consequently, the belief estimates returned by the semantic attachments guide the TRPG in identifying the \textit{helpful actions}, besides providing an efficient heuristic evaluation for the task plan.  

\newsavebox{\newlisting}
 \lstset{basicstyle=\small}
 \lstset{escapeinside={<@}{@>}} 
\begin{lrbox}{\newlisting}
\begin{lstlisting}

(<@\textcolor{cyan}{:durative-action}@> goto_region
 <@\textcolor{olive}{:parameters}@> (?v - robot ?from ?to - region)
 <@\textcolor{olive}{:duration}@> (= ?duration 100)
 <@\textcolor{olive}{:condition}@> (at start (robot_in ?v ?from))
 <@\textcolor{olive}{:effect}@> (and (at start (not (robot_in ?v ?from))) 
 (at start (increase (triggered ?from ?to) 1))
 (at end (robot_in ?v ?to)) (at end (assign (triggered ?from ?to) 0))	
 (at end (increase (act-cost) (external)))
 (at end (increase (goal-trace) (bound))))

(<@\textcolor{cyan}{:durative-action}@> collect_document
 <@\textcolor{olive}{:parameters}@> (?v - robot ?r - region)
 <@\textcolor{olive}{:duration}@> (= ?duration 20)
 <@\textcolor{olive}{:condition}@> (and (at start (robot_in ?v ?r)) (at start (> (get ?r) 0))
 (over all (robot_in ?v ?r)))
 <@\textcolor{olive}{:effect}@> (and (at end (collected ?r))(at end (increase (act-cost) 4))))

(<@\textcolor{cyan}{:durative-action}@> goto_lift
 <@\textcolor{olive}{:parameters}@> (?v - robot ?from ?to - region)
 <@\textcolor{olive}{:duration}@> (= ?duration 100)
 <@\textcolor{olive}{:condition}@>(and (at start (robot_in ?v ?from))
 (at start (=  (prepared) 1)))
 <@\textcolor{olive}{:effect}@> (and (at start (not (robot_in ?v ?from))) 
 (at start (increase (triggered ?from ?to) 1))
 (at end (reached ?to)) (at end (assign (triggered ?from ?to) 0))	
 (at end (increase (act-cost) (external))))

\end{lstlisting}
\end{lrbox}

\begin{figure}[t]

 \scalebox{0.93}{\usebox{\newlisting}}\hfill%
\caption{A fragment of the PDDL office domain.}
 \label{fig:domain}
\end{figure}
The feasibility for the motion plan $\tau_i^{j^*}$ is checked by accounting for the trace of the covariance matrix upon reaching a cubicle $s_{i+1}$, that is, $trace(\Sigma_{\tau_i^{j^*}(1)})$ . Since the cubicle doors are of specific length, we bound the trace by a constant $\eta$. However, the failure of an action $a_i$ to find a feasible motion plan during the current expansion doesn't mean that it has to be discarded. Feasibility also depends on the sequence of actions performed earlier. A different action sequence prior to $a_i$ can render $a_i$ feasible. Hence infeasible actions are not discarded and are set aside for reattempting later. Consequently the feasibility check is performed for the optimal plan returned (see lines 17-21).

\textit{Optimality:} For a given roadmap, the plan synthesized by our approach is optimal at the task-level. This means that the task plan cost returned by our approach ($\pi^*$) is lower than any of the other possible task plans ($\pi$). Let us consider that there exists a plan $\pi < \pi^*$. If both $\pi$ and $\pi^*$ have the same sequence of actions, this is not possible since the action costs are evaluated by the motion planner and for a given roadmap, the motion cost returned is the optimal for each action, giving $\pi^* \leq \pi$. If both $\pi$ and $\pi^*$ have a different sequence of actions, the task planner ensures that the returned plan is optimal, giving $\pi^* \leq \pi$. Therefore, in both the case, we have $\pi^* \leq \pi$.

\textit{Completeness:} We provide a sufficient condition under which the probability of our approach returning a plan approaches one exponentially with the number of samples used in the construction of the roadmap. A task planning problem, $\Omega = (S, A, \gamma, s_0, S_g)$ is complete if it does contain any dead-ends~\cite{hoffmann2001JAIR}, that is there are no states from which goal states cannot be reached. The PRM motion planner is probabilistically complete~\cite{karaman2011IJRR}, that is the probability of failure decays to zero exponentially with the number of samples used in the construction of the roadmap. Therefore, if the motion planner terminates each time it is invoked then probability of finding a plan, if it exists, approaches one. 

On the one hand our approach is probabilistically complete; on the other hand, it is also resolution complete since the motion plan feasibility depends on the parameter $\eta$. Nevertheless, given a fixed value of $\eta$, the probability that the planner fails to return a solution, if one exists, tends to zero as the number of samples approaches infinity. In this sense the best that we can guarantee is probabilistic completeness.

\section{Results}
\begin{figure}[]

  \subfloat[]{\includegraphics[scale=0.13]{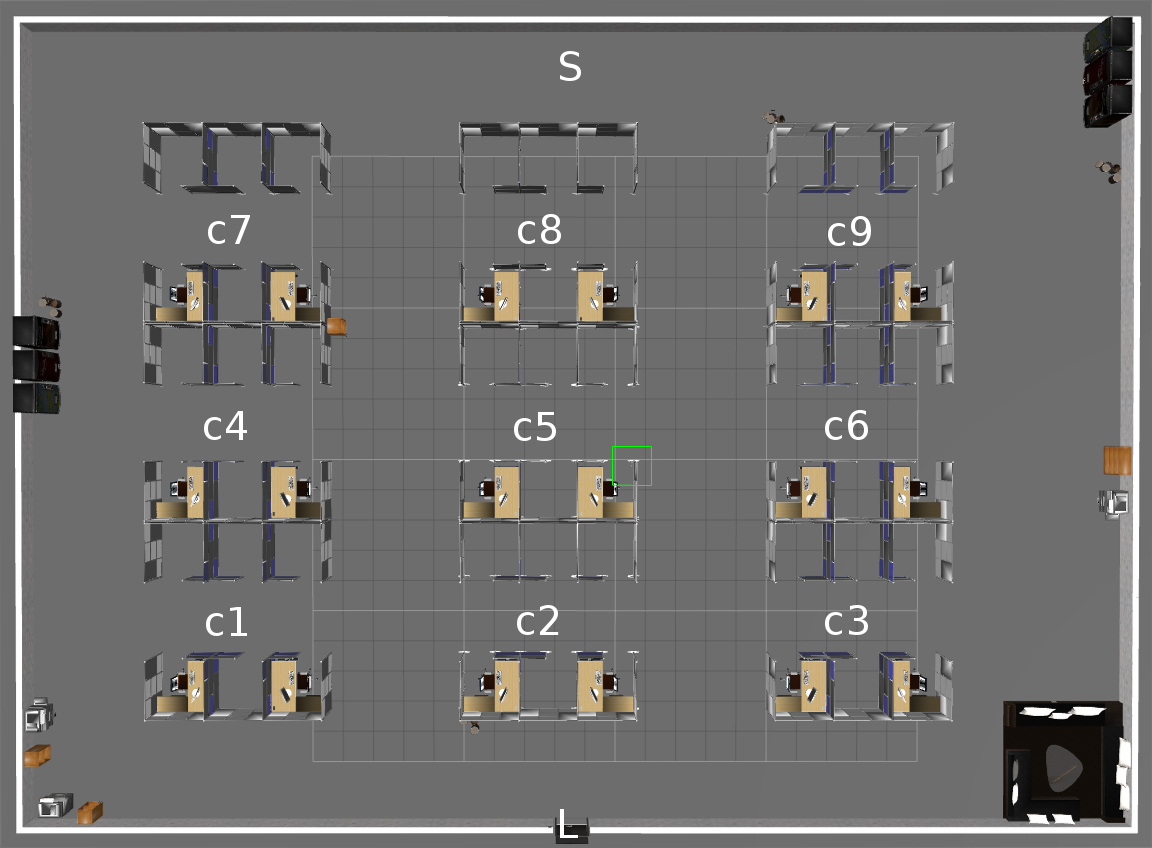}} \hspace{0.6cm}%
  \subfloat[]{\includegraphics[scale=0.355]{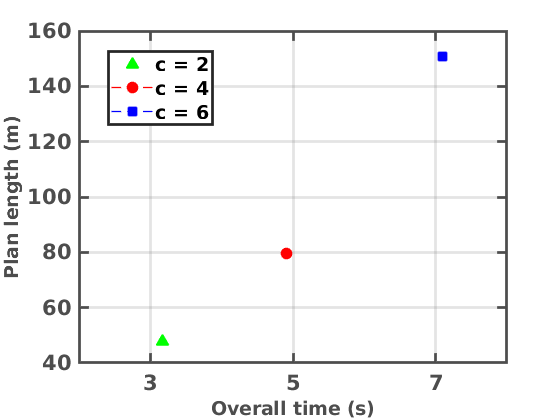}}
  \caption{(a) Simulated environment in Gazebo. See \textit{motivating example} in Section~\ref{sec:one} for a detailed description. (b) Plan length with overall planning time. \textit{config 1} is run with $d=1.5$.}
  \label{fig:gazebo_and_scale}
\end{figure}

To validate our approach we construct an office environment ($36m \times 25m$) in Gazebo as described in the \textit{motivating example} discussed in Section~\ref{sec:one}. The robot is required to collect documents from different cubicles which are then taken to the next floor via the lift $L$ (see Fig.~\ref{fig:office}). The top view of the simulated environment is shown in Fig.~\ref{fig:gazebo_and_scale}(a). We use the temporal POPF-TIF as our task planner by customizing it to dynamically load a shared library that performs a PRM based planning in the belief space. Though we use off-the-shelf task and motion planners, we would like to stress the fact that any task planner customized to perform semantic attachments can be employed for discrete planning. Our approach is also not restricted to roadmap based planners and can be easily adapted to different motion planners, for example RRT and its variants. The robot kinematics is modeled using the standard odometry based motion model. The performances are evaluated on an Intel{\small\textregistered} Core i7-6500U under Ubuntu 16.04 LTS.


\subsection{Validation}
\label{subsec:cost}
To benchmark our approach we consider four different configurations that differ in their motion cost computation. Though our formulation can be adapted to any general cost function, we choose the following four configurations to demonstrate the efficiency of our approach

\begin{itemize}
\item \textit{config 1}: In this configuration, the motion planner returns the trajectory length or the geometric-level cost of traversing from $s_i$ to $s_{i+1}$, that is $\tau_i(0) \in \phi(s_i)$ to $\tau_i(1) \in \phi(s_{i+1})$.
\item \textit{config 2}: Here the motion planner is never called and the task cost are evaluated computing the euclidean distance between the geometric instantiations of $s_i$ and $s_{i+1}$, that is, between $\tau_i^j(0)$ and $\tau_i^j(1)$.
\item \textit{config 3}: This configuration evaluates the motion cost as the sum of Euclidean distance between $\tau_i(0)$ and $\tau_i(1)$ and the cost due to uncertainty, defined as $c_{\Sigma}=trace(\Sigma)$, where $\Sigma$ is the covariance at each node of $\tau_i$. In addition, we also employ a term $c_{\Sigma_g}$, which is defined as the trace of the covariance at the goal node ($\tau_i(1) = \phi(s_{i+1})$) , where $s_{i+1}$ is a cubicle. The cubicle doors have a width of $2m$ and considering maximum uncertainty along the door width we fix $\eta = 1m$ as the maximum upper limit and discard the motion plans with $c_{\Sigma_g} > 1$.
\item \textit{config 4}: A combination of $c_{\Sigma}$ and $c_{\Sigma_g}$ is used for motion cost evaluation.
\end{itemize}


We first demonstrate the need for a combined TMP for navigation. Consider the following scenario in which th robot is required to collect documents from the cubicles $c3$, $c4$, $c6$ and $c9$. We first ran the planner with \textit{config 2} to synthesize the task plan. We remind that in this configuration, the motion planner is never called and the action costs are evaluated by considering the Euclidean distance between the start and goal regions. Essentially, \textit{config 2} correspond to planners that pre-compute motion costs of all task-level action. The plan synthesized is, $S\rightarrow c3 \rightarrow c4 \rightarrow c6 \rightarrow c9 \rightarrow L$.
This plan is then given to the motion planner, to compute the corresponding cost due to uncertainty $c_{\Sigma}$ and $c_{\Sigma_g}$. The task planning cost and the motion planning cost are added to estimate the overall planning cost, which equated to 298.84. In the same way, the overall planning time was computed to be 0.94 seconds. Next, we ran the planner with \textit{config 3}, returning the plan, $S\rightarrow c4 \rightarrow c9 \rightarrow c6 \rightarrow c3 \rightarrow L$, in 1.28 seconds with a total cost of 90.89. It is seen that there is a significant difference in the plan quality as the cost is improved by a factor of 3, clearly showing the efficiency of a combined TMP approach as opposed to performing them separately. This difference in cost is attributed to the different task sequence synthesized.

\begin{figure}[h!]
	\centering
	
\subfloat{\includegraphics[scale=0.1445]{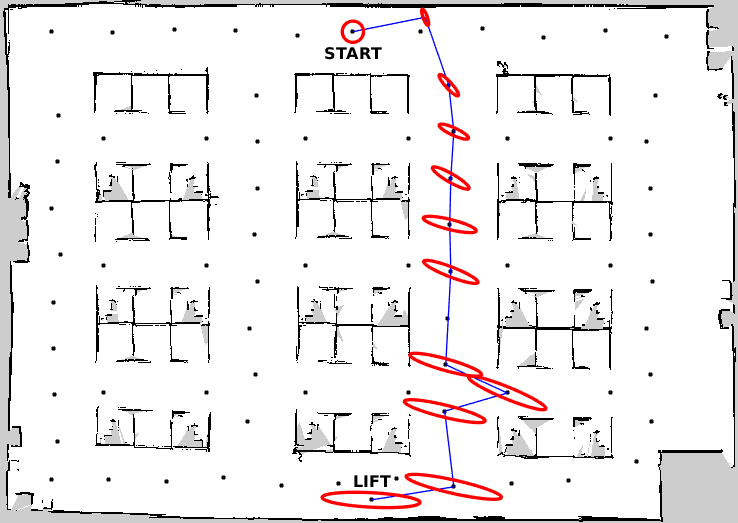}\label{fig:1}}
	\hspace{0.02cm}
\subfloat{\includegraphics[scale=0.145]{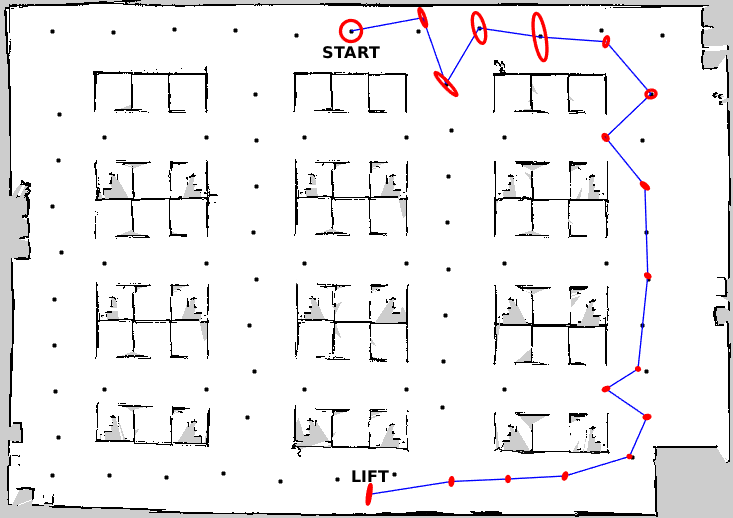}\label{fig:2}}
\hspace{0.02cm}
\subfloat{\includegraphics[scale=0.13]{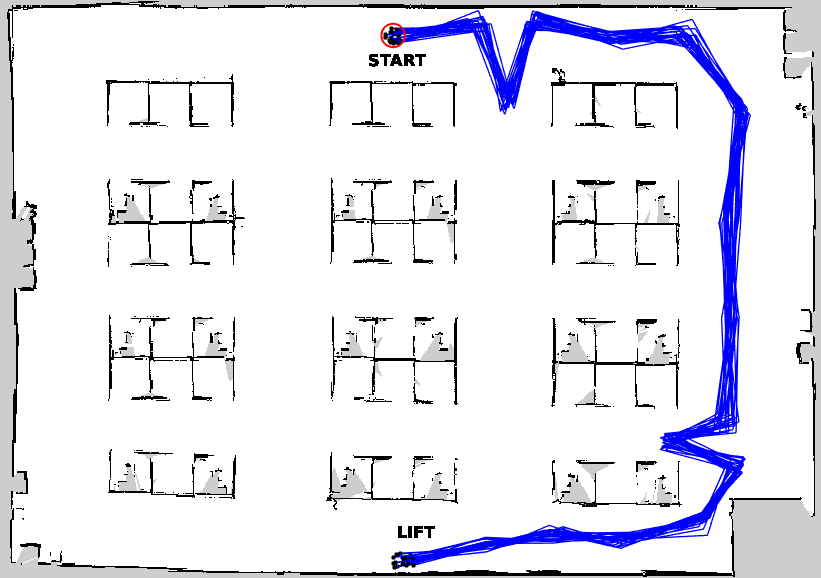}\label{fig:3}}
\vspace{0.05cm}
		\caption{(\textit{left} and \textit{center}) Figures showing the propagated belief distributions along the planned paths for \textit{config 1} and \textit{config 4}. The figures show the belief estimates for a single planning instantiation corresponding to a unique set of simulated observations. The dots (black), represents the sampled poses. (\textit{left}) Shortest path route that corresponds to \textit{config 1}. (\textit{center}) Belief space planning corresponding to \textit{config 4}, returning an information rich route. (right) Traces of robot's true state while starting from the initial belief.}
		\label{fig:BSP}
\end{figure}

Next, we run the planner with \textit{config 1} and \textit{config 4} to demonstrate the advantage of planning in belief space. Similar to PETLON~\cite{lo2018AAMAS}, in \textit{config 1}, the motion planner evaluates the geometric-level cost of traversing $\tau_i(0)$ to $\tau_i(1)$, whereas in \textit{config 4}, the cost due to uncertainty is returned. We consider a scenario in which the robot has to collect a document from cubicle $c3$. The planned trajectories in both the scenarios with the corresponding covariance estimated at each node (only the ($x$,$y$) portion shown) is shown in Fig.~\ref{fig:BSP}. Clearly, the belief space planner (\textit{config 4}) returns a route which is rich in sensor information (see center, Fig.~\ref{fig:BSP}), enabling effective localization. Fig.~\ref{fig:BSP} (right), shows the  traces of true robot state for 25 different simulations, the initial state being sampled from the known initial belief. 

\setlength{\tabcolsep}{8pt}
	\begin{table}
		{ 
			\begin{tabular}{l l  ccc | ccc }                             
			\hline 	
				\multicolumn{1}{c}{\multirow{ 1}{*}{Configuration}} &  \multicolumn{1}{c}{$d$} & \multicolumn{3}{c}{Overall time (s)} & \multicolumn{3}{c}{Cost}  \\
								{} & {} & \multicolumn{1}{c}{$c=2$} &  \multicolumn{1}{c}{$c=4$} & \multicolumn{1}{c}{$c=6$} & \multicolumn{1}{c}{$c=2$} &  \multicolumn{1}{c}{$c=4$} & \multicolumn{1}{c}{$c=6$} \\
				
				\hline        
\textit{config 1}	 & 1 & 0.47 & 0.77 & 1.77 & 47.80 & 84.88 & 161.47\\
                     & 1.5 & 3.17 & 4.91 & 7.10 & 55.77 & 95.74 & 174.90\\
                     & 2 & 6.08 & 9.86 & 15.14 & 56.19 & 95.77 & 181.06\\
                     \hline		
                     
                     {} & {} & \multicolumn{1}{c}{$c=2$} &  \multicolumn{1}{c}{$c=4$} & \multicolumn{1}{c}{$c=6$} & \multicolumn{1}{c}{$c=2$} &  \multicolumn{1}{c}{$c=4$} & \multicolumn{1}{c}{$c=6$} \\	
		       \hline              
\textit{config 4}	 &1 & 1.34 & 2.24 & - & 13.84 & 90.27 & -\\
                     & 1.5 & 3.41 & 7.16 & 14.04 & 20.18 & 57.01 & 79.86\\
                     & 2 & 9.11 & 28.48 & 46.15 & 16.32 & 54.96 & 82.02\\                     
                     \hline
		\end{tabular}}                                          
		\caption{Overall planning time and cost returned for \textit{config 1} and \textit{config 4} with different sample densities. $c$ = 2, 4 and 6 denotes the number of cubicles to be visited, increasing the task-level complexity.} 
\label{table:result1}
	\end{table}

\subsection{Scalability}
Finally, we test the scalability of our approach by increasing the task-level increasing the task-level complexity. We run our planner on three different scenarios where, 2, 4, 6 number of cubicles ($c=2,4,6$) are to be visited to collect the corresponding number of documents. This results in evaluating more task-level actions, escalating the task level complexity. We also test these scenarios on varying levels of sample densities. We choose $d=1,1.5,2$, where $d=i$ corresponds to an average of $i$ samples per square meter. The tests are run, using \textit{config 1} and \textit{config 4}. The overall planning time and the cost returned can be seen in Table~\ref{table:result1}. \textit{config 4} for $d = 1$ and $c=6$ did not produce a feasible motion plan as the condition $\eta < 1$ was violated. However, for higher sample densities, a feasible motion plan was found. The plan quality is increased with increase in $d$, but at the expense of exponentially increasing computation time. It is clearly seen that for our considered scenario $d=1.5$ can be chosen, without much loss of plan quality.

\begin{figure}[t]

  \subfloat[]{\includegraphics[scale=0.23]{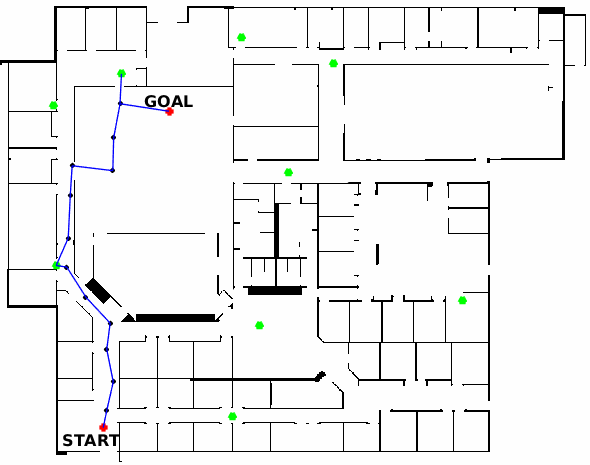}}\hfill%
  \subfloat[]{\includegraphics[scale=0.35]{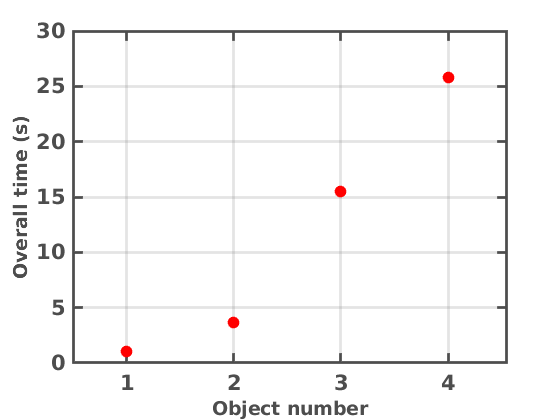}}
  \caption{(a) The optimal solution (blue path) for the willow garage environment when 2 documents are to be collected. The planner is run with \textit{config 1}. (b) Overall planning time with increasing number of objects to be collected for delivering.}
  \label{fig:willow}
\end{figure}

In ~\cite{lo2018AAMAS}, TMP for navigation is performed by evaluating the geometric cost of traversing and a scenario in which 2 objects are to be delivered to a person took about 15 seconds with a plan length of 37m. In comparison our approach fares superiorly with respect to increased task-level complexity. Though the environment considered in~\cite{lo2018AAMAS} is larger than ours, we evaluate the planning time with respect to the plan length by running our planner with \textit{conifg 1}. For a plan length of about 150m with $d=1.5$ and $c=6$, \textit{config 1} returned a plan in about 7 seconds (see Fig.~\ref{fig:gazebo_and_scale}(b)). To provide a better comparison, we also evaluate our approach by considering a much larger environment, the willow garage world ($58m \times 45m$) as shown in Fig.~\ref{fig:willow}(a). In this example, the robot needs to collect two objects (9 objects marked as green), to be delivered to a person at the goal location (shown in red). We ran our planner with \textit{config 1}, returning an optimal plan of length 53.94m in 3.69 seconds. This clearly elucidates the superiority of our approach. The scalability to increasing task complexity is tested by varying the number of objects to be collected (see Fig.~\ref{fig:willow}(b)). The task in which 4 objects are to be collected was completed in only about 25 seconds.

 The plan generated is executed with a TurtleBot robot in the simulated Gazebo environment. We use AprilTags~\cite{olson2011ICRA} to identify objects like printers, trash cans, as landmarks. The ROS infrastructure was used to perform the implementation. Belief estimation is carried out using EKF. 

\label{sec:results}

\section{Conclusion and Future Work}
This paper introduces an approach for task-motion planning under motion and sensing uncertainty. Task-motion interaction is facilitated by means of semantic attachments that return motion costs to the task planner. In this way, the action costs of the task plans are evaluated using a motion planner. The plan synthesized is optimal at the task-level since the overall action cost is less than that of other task plans generated for a given roadmap. The proposed approach is probabilistically complete and we have validated the framework using a simulated office environment in Gazebo. The approach has been evaluated with different configurations, clearly illustrating the need for a combined TMP approach for navigation in belief space. The results also suggest that our approach fares well with respect to increased task-complexity and plan length. 

For the motion planning search on the roadmap, we currently employ a breadth first search propagating beliefs along the edges and estimating the beliefs at each node using EKF. Breadth first search expands in a first-in, first-out fashion and given a roadmap, it does not guarantee a globally optimal solution. Hence, given a roadmap, the motion plan returned for each action is only locally optimal. A more judicious approach would involve employing an $\textrm{A}^\star$ search. However, developing an admissible heuristic for the search is challenging as the covariance evolution is not monotonic. Developing a suitable heuristic is a topic for future work. Yet, the plan synthesized at the task-level has an overall action cost that is less than that of other task plans generated for the given roadmap. It is to be noted that the action cost also encompasses the motion cost. Presently, as the number of samples vary, the search is performed again. It is our future direction to efficiently utilize the previous search results to reduce the computation time for increased samples. Currently, the planning is performed offline and we also plan to extend it to an online planner. 

\bibliographystyle{spphys.bst}
\bibliography{/home/antony/Research_Genoa/References/References}

\begin{thebibliography}{10}
\providecommand{\url}[1]{{#1}}
\providecommand{\urlprefix}{URL }
\expandafter\ifx\csname urlstyle\endcsname\relax
  \providecommand{\doi}[1]{DOI \discretionary{}{}{}#1}\else
  \providecommand{\doi}{DOI \discretionary{}{}{}\begingroup
  \urlstyle{rm}\Url}\fi

\bibitem{mcdermott1998AIPS}
D.~McDermott, M.~Ghallab, A.~Howe, C.~Knoblock, A.~Ram, M.~Veloso, D.~Weld,
  D.~Wilkins, in \emph{AIPS-98 Planning Competition Committee} (1998)

\bibitem{srivastava2014ICRA}
S.~Srivastava, E.~Fang, L.~Riano, R.~Chitnis, S.~Russell, P.~Abbeel, in
  \emph{Robotics and Automation (ICRA), IEEE International Conference on}
  (IEEE, 2014), pp. 639--646

\bibitem{kaelbling1998AI}
L.P. Kaelbling, M.L. Littman, A.R. Cassandra, Artificial Intelligence
  \textbf{101}(1-2), 99 (1998)

\bibitem{fikes1971strips}
R.E. Fikes, N.J. Nilsson, Artificial Intelligence \textbf{2}(3-4), 189 (1971)

\bibitem{nilsson1984shakey}
N.J. Nilsson, Shakey the robot.
\newblock Tech. Rep. 323, Airtificial Intellignece Center, {SRI} International,
  Menlo Park, California (1984)

\bibitem{dornhege2009SSRR}
C.~Dornhege, M.~Gissler, M.~Teschner, B.~Nebel, in \emph{Safety, Security \&
  Rescue Robotics (SSRR), IEEE International Workshop on} (IEEE, 2009), pp.
  1--6

\bibitem{cambon2009IJRR}
S.~Cambon, R.~Alami, F.~Gravot, The International Journal of Robotics Research
  \textbf{28}(1), 104 (2009)

\bibitem{hoffmann2003JAIR}
J.~Hoffmann, Journal of Artificial Intelligence Research \textbf{20}, 291
  (2003)

\bibitem{piacentini2015AI}
C.~Piacentini, V.~Alimisis, M.~Fox, D.~Long, Artificial intelligence
  \textbf{229}, 210 (2015)

\bibitem{kaelbling2012aTR}
L.P. Kaelbling, T.~Lozano-P{\'e}rez, Integrated robot task and motion planning
  in the now.
\newblock Tech. Rep. 2012-018, Computer Science and Artificial Intelligence
  Laboratory, Massachusetts Institute of Technology (2012)

\bibitem{kaelbling2013IJRR}
L.P. Kaelbling, T.~Lozano-P{\'e}rez, The International Journal of Robotics
  Research \textbf{32}(9-10), 1194 (2013)

\bibitem{garrett2018IJRR}
C.R. Garrett, T.~Lozano-Perez, L.P. Kaelbling, The International Journal of
  Robotics Research \textbf{37}(1), 104 (2018)

\bibitem{toussaint2015IJCAI}
M.~Toussaint, in \emph{Twenty-Fourth International Joint Conference on
  Artificial Intelligence} (2015)

\bibitem{lozano2014IROS}
T.~Lozano-P{\'e}rez, L.P. Kaelbling, in \emph{Intelligent Robots and Systems
  (IROS), IEEE/RSJ International Conference on} (IEEE, 2014), pp. 3684--3691

\bibitem{dantam2018IJRR}
N.T. Dantam, Z.K. Kingston, S.~Chaudhuri, L.E. Kavraki, International Journal
  of Robotics Research, Special Issue on the 2016 Robotics: Science and Systems
  Conference \textbf{37}(10), 1134 (2018)

\bibitem{dornhege2009ICAPS}
C.~Dornhege, P.~Eyerich, T.~Keller, S.~Tr{\"u}g, M.~Brenner, B.~Nebel, in
  \emph{International Conference on Automated Planning and Scheduling (ICAPS)}
  (Thessaloniki, Greece, 2009), pp. 114--121

\bibitem{lo2018AAMAS}
S.Y. Lo, S.~Zhang, P.~Stone, in \emph{Proceedings of the 17th International
  Conference on Autonomous Agents and MultiAgent Systems} (International
  Foundation for Autonomous Agents and Multiagent Systems, 2018), pp. 220--228

\bibitem{ghallab2004automated}
M.~Ghallab, D.~Nau, P.~Traverso, \emph{{Automated Planning: Theory and
  Practice}} (Elsevier, 2004)

\bibitem{fox2003JAIR}
M.~Fox, D.~Long, Journal of artificial intelligence research \textbf{20}, 61
  (2003)

\bibitem{weyhrauch1980AI}
R.W. Weyhrauch, Artificial Intelligence \textbf{13} (1980)

\bibitem{bernardini2017ICAPS}
S.~Bernardini, M.~Fox, D.~Long, C.~Piacentini, in \emph{International
  Conference on Automated Planning and Scheduling (ICAPS)} (Pittsburgh, PA,
  USA, 2017), pp. 29--37

\bibitem{coles2010ICAPS}
A.J. Coles, A.I. Coles, M.~Fox, D.~Long, in \emph{Twentieth International
  Conference on Automated Planning and Scheduling} (2010)

\bibitem{kavraki1996IEEE}
L.E. Kavraki, P.~Svestka, J.C. Latombe, M.H. Overmars, IEEE transactions on
  Robotics and Automation \textbf{12}(4), 566 (1996)

\bibitem{hoffmann2001JAIR}
J.~Hoffmann, B.~Nebel, Journal of Artificial Intelligence Research \textbf{14},
  253 (2001)

\bibitem{karaman2011IJRR}
S.~Karaman, E.~Frazzoli, The international journal of robotics research
  \textbf{30}(7), 846 (2011)

\bibitem{olson2011ICRA}
E.~Olson, in \emph{Proceedings of the {IEEE} International Conference on
  Robotics and Automation ({ICRA})} (IEEE, 2011), pp. 3400--3407

\end{thebibliography}
\end{document}